\documentclass{bmvc2k}

\title{Highly Efficient SNNs for High-speed Object Detection}
\usepackage[T1]{fontenc}
\usepackage{graphicx}
\usepackage{graphicx}
\usepackage{amsmath}
\usepackage{amssymb}
\usepackage{booktabs}
\usepackage{subfigure}
\usepackage{soul}
\usepackage{changes}
\addauthor{Nemin Qiu}{qiunm@bupt.edu.cn}{1}
\addauthor{Zhiguo Li}{youxianglizhiguo@163.com}{2}
\addauthor{Yuan Li}{yuanli@pku.edu.cn}{3}
\addauthor{Chuang Zhu}{czhu@bupt.edu.cn}{*}
 
\addinstitution{
 School of Artificial Intelligence,\\
 Beijing University of Posts and Telecommunications\\
 Beijing, China
}
\addinstitution{
 Peking University\\
 Beijing, China
}

\runninghead{Student, Prof, Collaborator}{BMVC Author Guidelines}

\begin{document}

\maketitle

\begin{abstract}
The high biological properties and low energy consumption of Spiking Neural Networks (SNNs) have brought much attention in recent years. However, the converted SNNs generally need large time steps to achieve satisfactory performance, which will result in high inference latency and computational resources increase. In this work, we propose a highly efficient and fast SNN for object detection. First, we build an initial compact ANN by using quantization training method of convolution layer fold batch normalization layer and neural network modification. Second, we theoretically analyze how to obtain the low complexity SNN correctly. Then, we propose a scale-aware pseudo-quantization scheme to guarantee the correctness of the compact ANN to SNN. Third, we propose a continuous inference scheme by using a Feed-Forward Integrate-and-Fire (FewdIF) neuron to realize high-speed object detection. Experimental results show that our efficient SNN can achieve 118$\times$ speedup on GPU with only 1.5MB parameters for object detection tasks. We further verify our SNN on FPGA platform and the proposed model can achieve 800+FPS object detection with extremely low latency.
\end{abstract}

\section{Introduction}
\label{sec:intro}
\renewcommand{\thefootnote}{}
\footnotetext{Corner symbol ’*’ in the author name means the corresponding author.}
\begin{figure}
  \centering
  \includegraphics[width=0.98\textwidth]{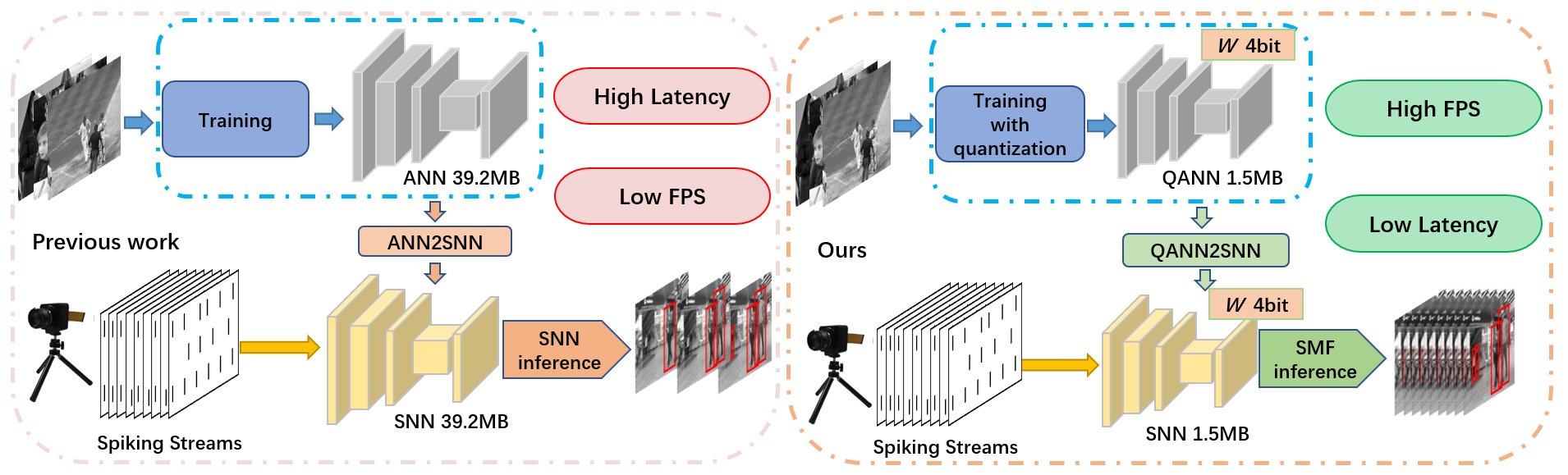}
  \caption{Illustration of our method and the previous work for SNN object detection. Previous work get the SNN model (39.2MB) by converting trained ANN with 32-bit precision\cite{rueckauer2017conversion,kim2020spiking} (ANN2SNN). It causes high latency and low FPS. Our method first obtains an initial quantized ANN (QANN), and then obtains a model size of only 1.5MB SNN with 4-bit precision by QANN2SNN conversion. FPS can be further increased using our proposed SNN continuous inference. We deploy and implement 800+FPS detection on FPGA using only 2.4W of power.}
  \label{introduction}

\end{figure}

 Artificial Neural Networks  (ANNs) have achieved great success in computer vision~\cite{redmon2016you}, natural language processing\cite{vaswani2017attention} and other fields. However, the success of ANNs is also accompanied by some serious concerns on their huge demand on computational resources and power consumption. In contrast, the human brain can provide excellent cognitive abilities with ultra-low natural power. Thus, many brain-inspired Spiking Neural Networks (SNNs)\cite{ding2021optimal,wu2019direct} are proposed to decrease computational resources and power consumption. SNNs are viewed as the third generation of neural network models, using biologically-realistic but simplified models of neurons to carry out computation. The event-driven mechanism in SNNs greatly avoids consuming excessive resources to a large extent\cite{kim2020spiking}. SNNs are suitable to be implemented on low-power mobile or edge devices \cite{kim2022exploring,liu2022dynsnn}. 

 At present, direct training SNNs and ANN to SNN conversion are two ways to generate SNN model. The SNN model obtained by direct training suffers unsatisfactory accuracy\cite{wang2022signed} due to the use of surrogate gradient\cite{wu2019direct,shrestha2018slayer,tavanaei2019bp} to address the non-differentiable binary activation function. The converted SNNs can obtain satisfactory performance, and we focus on this kind of SNN model in this paper. However, to maintain decent model precision, the converted SNNs generally need large time steps (such as  work\cite{kim2020spiking}  taking thousands of time steps in object detection), which result in high inference latency\cite{roy2019towards,wang2022signed,li2022spike} and computational resources increase\cite{roy2019towards,wang2022efficient}. Moreover, the converted SNNs still suffer large model size due to the corresponding high complex ANNs. Fig. \ref{introduction} illustrates the large SNN model of the previous works\cite{rueckauer2017conversion,kim2020spiking} need to accumulate many time steps to achieve decent performance, which result in low FPS (Frames Per Second).


In this work, we propose a highly efficient and fast SNN for object detection. 
First, we build an initial compact ANN by using quantization training method of convolution layer fold batch normalization layer and neural network modification.
Second, we theoretically analyze how to obtain the low complexity SNN correctly by using conversion method. Meanwhile, we propose a
scale-aware pseudo-quantization scheme to guarantee the correctness of the quantized ANN to SNN. Then we obtain a highly efficient low complexity SNN. 
Third, we propose a continuous inference scheme to realize high-speed object detection. Specifically, to support our continuous inference, we design a Feed-Forward Integrate-and-Fire (FewdIF) neuron which is capable of accumulating history information.

To summarize, our main contributions are as follows:


\begin{itemize}
\item We propose a highly efficient and fast SNN for object detection. Specifically, we first convert the quantized ANN to low complexity SNN, and then construct a continuous inference scheme to realize high-speed object detection.

\item In the SNN conversion, we first perform quantization training method of convolution layer fold batch normalization layer and neural network modification. Then, we propose a scale-aware pseudo-quantization scheme to guarantee the correctness of the quantized ANN to SNN. 

\item In the inference stage, we propose a continuous inference scheme to realize high-speed object detection by using our designed FewdIF neuron.

\item Experimental results show that our efficient SNNs have few and low bit-width parameters (1.5MB) and high-speed detection (GPU: FPS 177.5 vs 1.5\cite{kim2020spiking}) on object detection tasks. We further deploy the SNNs on FPGA and achieve 800+FPS detection with extremely low latency.
\end{itemize}




\section{Related Work}
\textbf{ANN to SNN Conversion}: The conversion of ANN to SNN is in burgeoning research. Cao $et$ $al$. \cite{cao2015spiking} proposed a ANN to SNN conversion method that neglected bias and max-pooling. In the next work, Rueckauer $et$ $al$.\cite{rueckauer2017conversion} presented an implementation method of batch normalization and spike max-pooling. Meanwhile, To get deeper SNNs. Diehl $et$ $al$.\cite{diehl2015fast} proposed the data-based normalization to improve the performance in deep SNNs. Sengupta $et$ $al$.\cite{sengupta2019going} expanded conversion methods to VGG and residual architectures. However, the converted SNN requires massive time steps to reach competitive performance\cite{roy2019towards}. All of them are complicated procedures vulnerable to high inference latency\cite{wang2022signed,park2019fast}. To reduce the time step, Park $et$ $al$.\cite{park2019fast} proposed a fast and energy-efficient information transmission method with burst spikes and hybrid neural coding scheme in deep SNNs. Ding $et$ $al$.\cite{ding2021optimal} presented Rate Norm Layer to replace the ReLU function, and obtain the scale through a gradient-based algorithm. Nonetheless, most previous works have been limited to the image classification task. 

\textbf{Object Detection for SNN}: Kim $et$ $al$.\cite{kim2020spiking} have presented Spiking-YOLO, the first SNN model that successfully performs object detection by achieving comparable results to those of the original ANNs on non-trivial datasets, PASCAL VOC and MS COCO. However, it suffers from high inference latency and computational resources increase\cite{wang2022signed}. Moreover, the converted SNNs still suffer large model size due to the corresponding high complex ANNs.

 \textbf{Model Compression}: In the field of pruning neural networks, the pruning methods \cite{liu2017learning,ye2018rethinking} usually compresses the model and accelerates the inference. In the field of of quantization. Jacob $et$ $al$.\cite{jacob2018quantization} propose a quantization scheme that relies only on integer arithmetic to approximate the floating-point computations in a neural network.

There are no efforts to compress and accelerate SNNs on detection tasks. In this paper, based on the existing work, we further design  a highly efficient and fast SNN for object detection.

\section{Method}
\begin{figure*}
  \centering
  \includegraphics[width=1\textwidth]{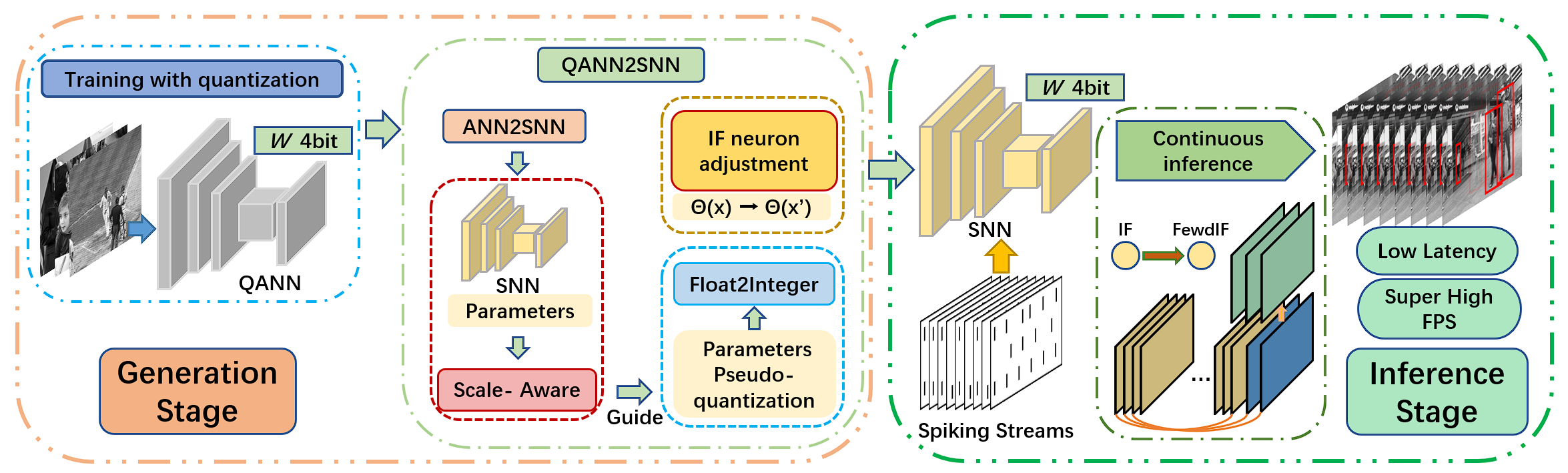}
  \caption{The Illustration for achieving super high FPS and low latency object detection by using the proposed highly efficient SNN. There are two stages. In generation stage, we build an initial compact quantized ANN (QANN) by using quantization training method of convolution layer fold batch normalization layer and neural network modification. Then we correctly convert the quantized ANN to SNN (QANN2SNN) by using our proposed scale-aware pseudo-quantization scheme. Thus we get a highly efficient low complexity SNN. In inference stage, replacing the original IF neurons with proposed FewdIF neurons can achieve the proposed SNN continuous inference which can greatly improve FPS.}
  \label{fig_sys_all}
\end{figure*}

In this section, we will present how we implement the highly efficient and fast SNN from two stages, generation and inference, respectively. Fig. \ref{fig_sys_all} is the overview of the generation stage and inference stage. The generation stage contains the quantization training of the initial compact ANN (QANN) in Section \ref{section4.1} and the conversion of the quantized ANN to the low complexity SNN (QANN2SNN) in Section \ref{section4.2}. The inference stage includes Feed-Forward Integrate-and-Fire (FewdIF) neurons and SNN continuous inference we proposed in Section \ref{section4.3}.

\subsection{ANN Quantization}\label{section4.1}

In this section, we focus on the preparing work for efficient SNN generation. We build an initial compact ANN by using quantization training method of convolution layer fold batch normalization layer and neural network modification.

 Specifically, we reduce the bit-width of the ANN weights by using quantization. The low bit-width compact ANNs can be correctly converted to SNN by using QANN2SNN method in Section \ref{section4.2}. This allows the weight bit-width of the converted SNN to be further reduced as well. Considering that the performance of the converted SNN depends on its initial ANN \cite{panchapakesan2022syncnn}, we need to build an initial compact ANN that performs well and is suitable for conversion to SNN.



We build the initial compact quantized ANN by using the training in the Fig. \ref{quantized ANN generation}. It is a training method adapted for QANN2SNN method. In particular, we generate initial compact quantized ANN by using quantization training method of convolution layer fold batch normalization layer and neural network modification. Noteworthy, to obtain better ANN, we use training with simulated quantization \cite{jacob2018quantization,krishnamoorthi2018quantizing} as our method of quantization training. What's more, we train with Quantized ReLU (QReLU) instead of ReLU, which not only completes the operation of quantizing activation values, but also reduces the time steps of SNN \cite{ding2021optimal}. For better SNN performance after conversion, we use down-sampling convolution to replace max-pooling layer. The upsampling layer in ANN is replaced by transpose convolution.

\subsection{Quantized ANN to SNN}\label{section4.2}


In order to avoid the bit-width rise of SNN weights after the conversion, we propose a scale-aware pseudo-quantization scheme to guarantee the correctness of the quantized ANN (QANN) to SNN. The conversion of QANN to SNN consists of the following three steps: weight conversion, weight bit-width mapping and type conversion, and IF neuron adjustment. To simplify the description, in this section, let $Q( \cdot )$ denotes the int8 quantization function.

\textbf{ANN to SNN}: 
The similarity of Integrate-and-Fire (IF) neuron and ReLU activation functions\cite{rueckauer2017conversion} is an important basis on which ANNs can be converted to SNNs. The principle of ANN to SNN conversion is that the firing rates of spiking neuron $r_{k}^{l}(T)$ should correlate with the original ANN activations $x_{k}^{l}$ such that $r_{k}^{l}(T){\rightarrow} x_{k}^{l}$. The firing rate of each SNN neuron as $r_{k}^{l}(T)=N_{k}^{l}(T)/T$, where $N_{k}^{l}(T)=\sum_{t=1}^{T}\,\Theta_{t,k}^{l}$ is the number of spikes generated in $T$ time steps, let's $\Theta_{t,k}^{l}$ denotes a step function indicating the occurrence of a spike at time $t$. For activation function mapping  $r_{k}^{l}(T){\rightarrow} x_{k}^{l}$ in ANN to SNN conversion, there has been a lot of previous works\cite{rueckauer2017conversion,ding2021optimal,kim2020spiking} that describes this (Theory for Conversion from ANN to SNN) in detail. The layer-to-layer relationship between the firing rates of IF neurons obtained through a series of derivations and approximations is:
\begin{equation}
  r_{k}^{l}(T)\approx\sum_{j}(w_{k,j}^{l}\cdot {r_{j}^{l-1}(T)})+b\sb{k}\sp{l}.
  \label{eq:r3}
\end{equation}
  \vskip -0.1 in
This relationship is very similar to the ANN's layer-to-layer activation value relationship: 

\begin{equation}
 x_{k}^{l}=\sum_{j}(w\sp{l}\sb{k,j}\cdot\,x\sb{j}^{l-1})+b\sb{k}\sp{l}.
  \label{eq:r4}
\end{equation}
  \vskip -0.1 in
\textbf{Weight conversion}: From the above definition, it is clear that the firing rate of IF neurons $r_{k}^{l}(T)\,\in\,[0,1]$, we need to adjust the output range of ANNs ReLU activation function to [0, 1]\cite{rueckauer2017conversion}. Therefore, we achieve this adjustment by converting the parameters of the ANNs. The well-known layer-wise parameter normalization (LayerNorm)\cite{rueckauer2017conversion} is a typical parameters transformation method. Specifically, after the ANN model is trained, we need to count the input tensor and output tensor of this layer. The maximum value of the input tensor is $M^{l-1}$, the maximum value of the output tensor is $M^{l}$, and the normalized weight and bias should be as follow:
 \begin{equation}
  \hat{w}_{k,j}^{l}=\frac{w_{k,j}^{l}\cdot M^{l-1}}{M^{l}},\hspace{0.5cm}
  \hat{b}_{k}^{l}={\frac{b_{k}^{l}}{M^{l}}}.
  \label{eq:otherlayer_layernorm}
\end{equation}
where $w\sp{l}\sb{k,j}$ represents weights, $b\sb{k}\sp{l}$ represents biases. After completing the above operations, replace the ReLU activation function in the ANN with the IF neuron. 

In order to get the SNNs with low bit-width parameters, let us introduce the equation for quantized ANNs $x\sb{k}^{l}$ as shown in Eq. (\ref{eq:quant_activation_ANN}).
\begin{equation}
  Q(x\sb{k}^{l})=f(\sum_{j=0}\sp{n}(Q(w\sp{l}\sb{k,j})\cdot\,x\sb{j}^{l-1})+Q(b\sb{k}\sp{l})).
  \label{eq:quant_activation_ANN}
\end{equation} 

By using the previous ANN to SNN conversion method, some adjustments are made to the Eq. (\ref{eq:otherlayer_layernorm}). The maximum value of the input tensor after quantization is $Q(M^{l-1})$, the maximum value of the output tensor after quantization is $Q(M^{l})$, and the normalized weights and biases should be as follow:
\begin{equation}
  \hat{w}_{k,j}^{l}=\frac{Q(w_{k,j}^{l})\cdot Q(M^{l-1})}{Q(M^{l})},\hspace{0.5cm}
  \hat{b}_{k}^{l}={\frac{Q(b_{k}^{l})}{Q(M^{l})}}.
  \label{eq:secondcase_w_and_b}
\end{equation}

 We find a problem encountered in converting the quantized ANN to SNN according to Eq. (\ref{eq:secondcase_w_and_b}). Specifically, the converted weights $\hat{w}_{k,j}^{l}$ and biases ${\hat{b}_{k}^{l}}$ are obtained by multiplying corresponding int8 values according to Eq. (\ref{eq:secondcase_w_and_b}). The bit-width of parameters is obviously increased, because two operations with high bit numbers require higher bits to store lossless results. 

\textbf{Weight bit-width mapping and type conversion}: To solve the above problems, we propose a scale-aware pseudo-quantization scheme to guarantee the correctness of the quantized ANN to SNN. We divide them by the minimum interval of two numbers, and these parameters can still be stored with int8 bit-width. Let $U_{k}^{l}(t)$ denote a transient membrane potential increment of spiking neuron $k$ in layer $l$, Our method uses $\hat{U}_{k}^{l}(t)$ instead of $U_{k}^{l}(t)$:
\begin{equation}
  \hat{U}_{k}^{l}(t)=\sum_{j}(\hat{w}_{k,j}^{l}\cdot S_{l} \cdot \Theta_{t,j}^{l-1})+\hat{b}\sb{k}\sp{l}\cdot S_{l} =U_{k}^{l}(t)\cdot S_{l} ,
  \label{eq:U_sacle_up}
\end{equation}
lets $s_{l}$ denotes the minimum of the absolute value of the difference between any of the weight values in $l$ layer. Where $S_{l}$ represents $S_{l}=1/s_{l}$. $\Theta_{t,k}^{l}$ denotes the output of the spiking neuron $k$ at moment $t$. By type conversion we can get the integer $Int(w_{k,j})$:
\begin{equation}
  Int(w_{k,j})=Int(\hat{w}_{k,j}^{l}\cdot S_{l}).
  \label{eq:int_W}
\end{equation}

$Int(w_{k,j})$ will be used in the inference. For the biases, according to Eq. (\ref{eq:secondcase_w_and_b}), it is known that the minimum interval of the bias is not necessarily the same as the minimum interval of the weights. Therefore, for the biases, we use  32-bit floating point storage or direct rounding to 32-bit int type. Although the biases are quantized as 32-bit values, they account for only a tiny fraction of the parameters in a neural network \cite{jacob2018quantization}.

\textbf{IF neuron adjustment}: According to the definition of spiking neuron output $\Theta_{t,j}^{l}$, the spiking neuron integrates inputs $U_{k}^{l}(t)$ until the membrane potential $V_{k}^{l}(t-1)$ exceeds a threshold $V_{k,th}$ and a spike is generated. In the case of using our method, if we want to ensure that the output of  $\Theta_{t,j}^{l}$ is not affected by linear change $\hat{U}_{k}^{l}(t)$, we need to multiply $V_{k,th}$ also with the scale factor $S_{l}$:
\begin{equation}
  \Theta_{t,k}^{l}=\Theta(V_{k}^{l}(t-1)+\frac{(\hat{U}_{k}^{l}(t)-S_{l}\cdot V_{k,th})}{S_{l}}).
  \label{eq:NEW_Theta}
\end{equation}

\begin{figure}
  \centering
  \subfigure[Train]{\includegraphics[width=0.46\textwidth]{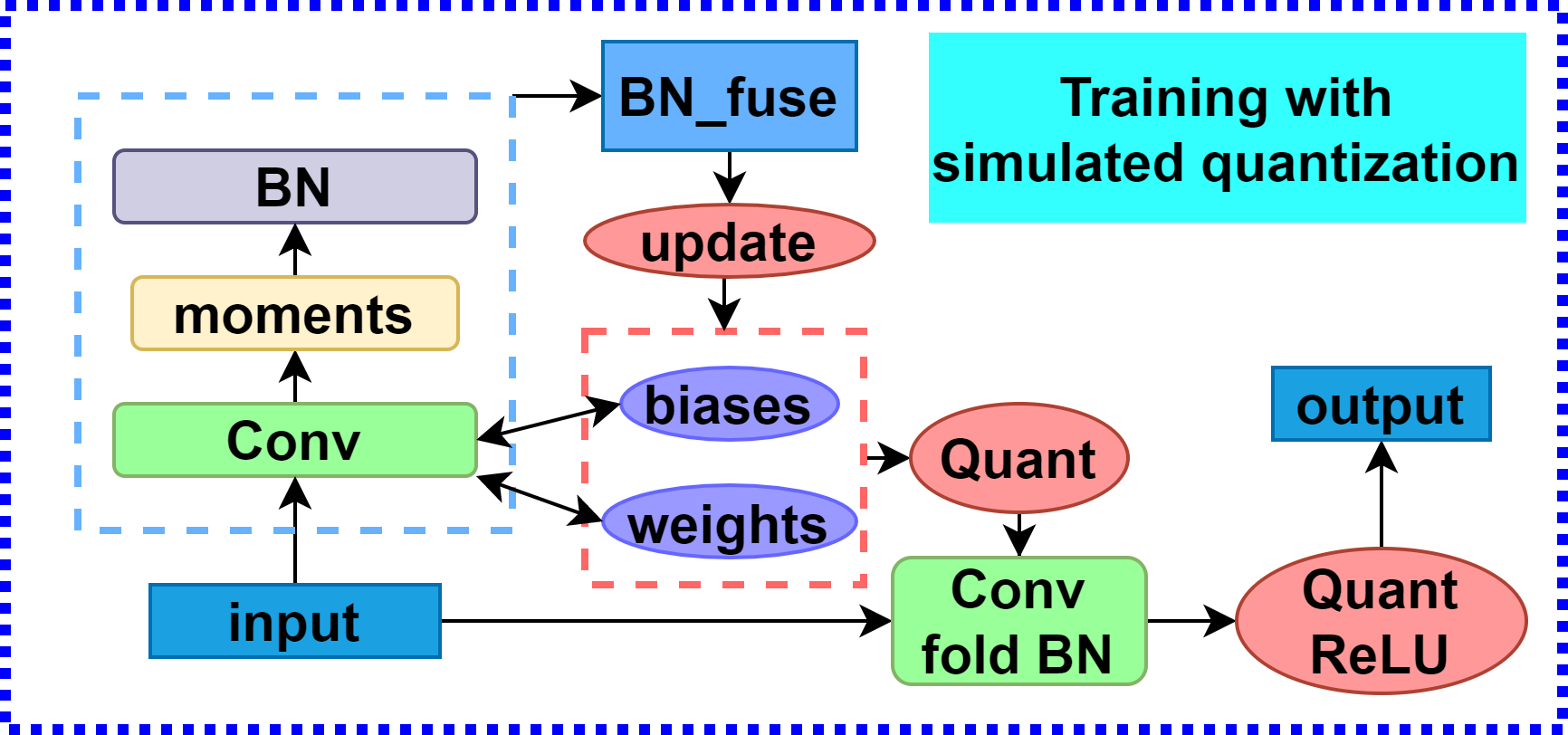}}
  \subfigure[Infer]{\includegraphics[width=0.46\textwidth]{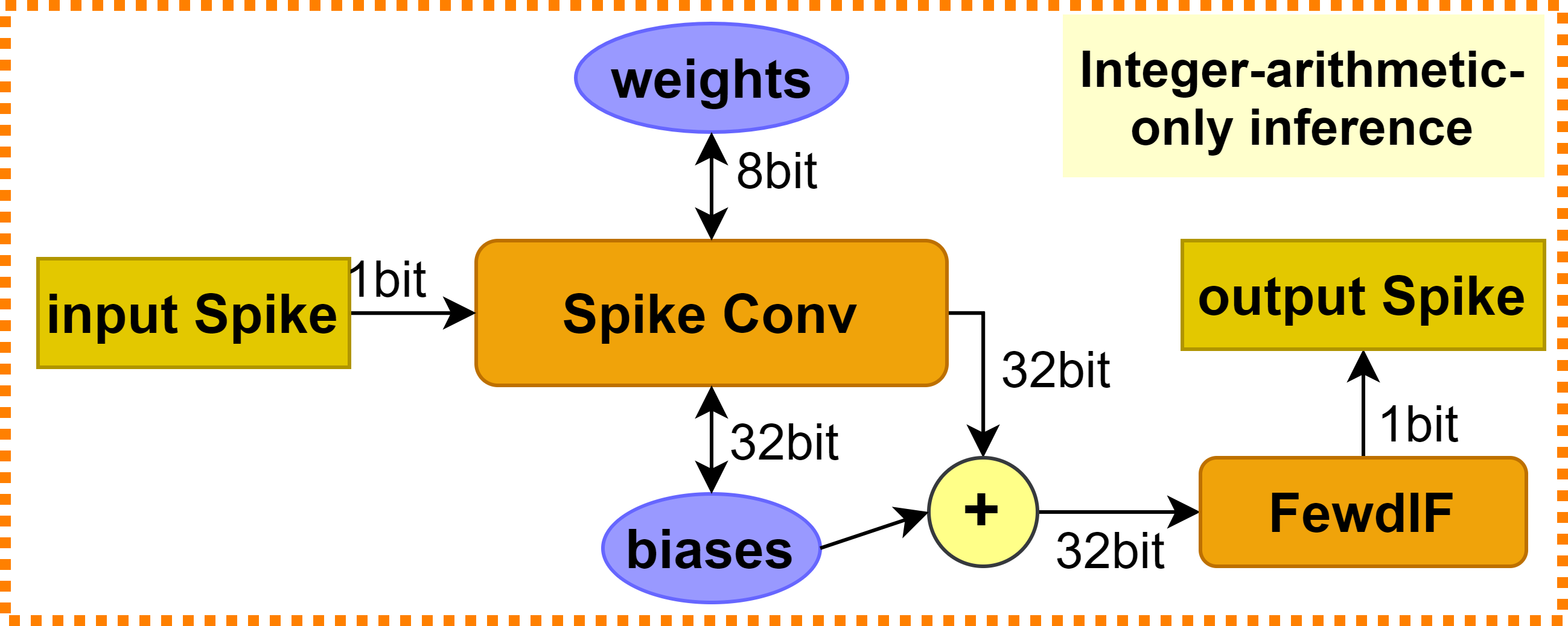}}
  \caption{Quantization training of initial compact ANN and Integer-arithmetic-only inference of the low complexity SNN.}
  \label{quantized ANN generation}
  \label{integer_arithmetic_infer}
\end{figure}

After finishing these corrections, we successfully solve the problem of converting a low bit-width ANN to a low bit-width SNN. Finally, we can use the SNN Integer-arithmetic-only inference architecture shown in Fig. \ref{integer_arithmetic_infer}. Experimental results show that our efficient SNNs with few and low bit-width parameters overcome high latency on object detection tasks. Compared to previous methods, our SNN model with 4-bit parameters exceeds the performance of previous methods using few time steps.  
  \vskip -0.1 in

\subsection{SNN Continuous Inference}\label{section4.3}

\begin{figure}
  \label{SCI}
  \centering
  \subfigure[SNN Inference]{\includegraphics[width=0.43\textwidth]{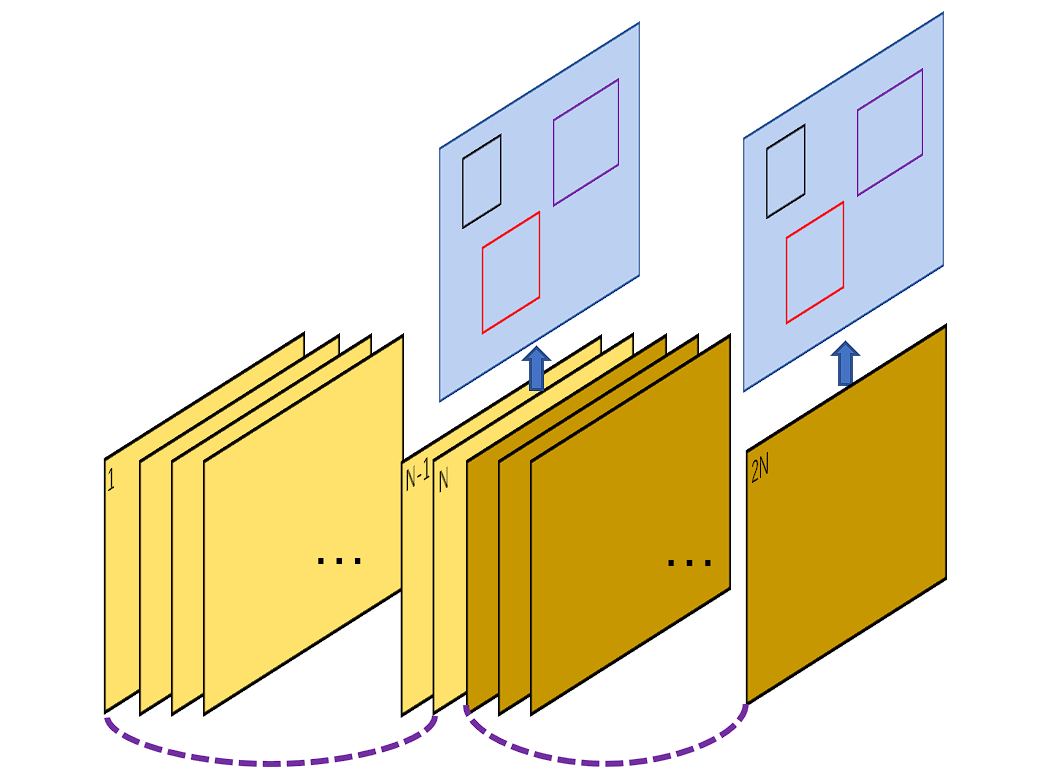}}
  \subfigure[SNN Continuous Inference]{\includegraphics[width=0.32\textwidth]{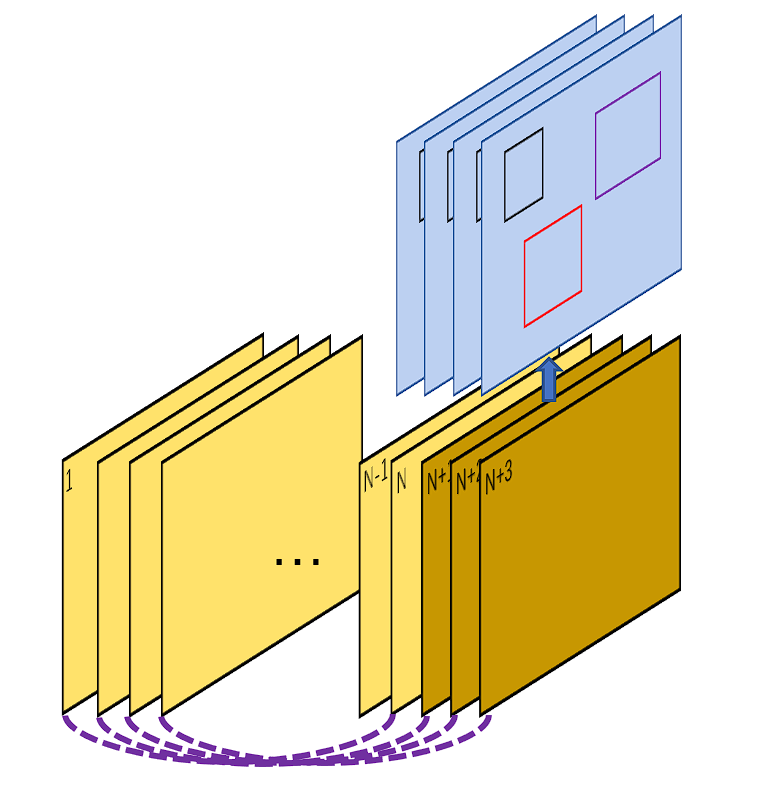}}
  \caption{Comparison of (a) SNN inference and (b) Continuous inference. With the same number of spiking datas, more results are output using continuous inference.}
  \label{SMF}
\end{figure}

Most of the previous ANN to SNN works focus on single image tasks. Their SNN inference \cite{kim2020spiking,rueckauer2017conversion} is shown in Figure \ref{SMF} (a). SNN inferring one frame result need to accumulate N frames of spike data each time to correspond to the output one frame result as ANN. The neuronal membrane potential of IF neuron is reset to 0 after every N time steps. Considering the continuous scenario, we believe that such an inference approach does not make good use of the spiking data. We propose a continuous inference scheme shown in Figure \ref{SMF} (b). We do not set the membrane potential to 0 in the continuous scenario. In this way, the first frame of the SNN output needs the input of the spike frames from the first frame to $N_{th}$ frame. While the second only needs the input of new $(N+1)_{th}$ spike frame to predict a result. It is different from the previous inference which needs the input of $(N+1)_{th}$ frame to $2N_{th}$ frame. However, IF neurons use this method with severe performance degradation.

To solve the above problems, we propose Feed-Forward Integrate-and-Fire (FewdIF) neurons to avoid excessive “excitation” and “inhibition” of IF neurons. The purpose of the modifications is to limit the maximum and minimum accumulation of membrane potentials to ensure that the previous frame may affect the input of the next frame, but not “overcall”. The positive and negative boundary values of the two membrane potentials that a neuron's membrane potential can accumulate to at most is defined as follows:
\begin{equation}
  MAX(V_{k}^{l}(t))_{FewdIF}=(N_{max} \cdot V_{k,th}),
  \label{MAX_FewdIF_V}
\end{equation}
\begin{equation}
  MIN(V_{k}^{l}(t))_{FewdIF}=(N_{min} \cdot V_{k,th}),
  \label{MIN_FewdIF_V}
\end{equation}
where $N_{max}$ is the maximum scale factor and $N_{min}$ is the minimum scale factor of FewdIF neuron. By using FewdIF neurons instead of IF neurons in an SNN, continuous inference can be achieved with a single time step after the SNN has adapted to the scenario. 

Experiments show that the SNN continuous inference only need one time step to predict. Compared to the previous work on object detection \cite{kim2020spiking}, we significantly reduce the time steps.

\section{Experiments}
Since there are almost no researches in this area yet, we did our best to conduct the comparison with relevant experiments\cite{kim2020spiking,li2022spike,nguyen2019high,ahmad2020accelerating}.
In the Section \ref{section5.1}, we set up a performance comparison of two SNNs on object detection task to validate our low complexity SNN obtained in Section \ref{section4.2}. In the Section \ref{section5.2}, we set up a comprehensive comparison of different SNN inference methods to verify our FewdIF neuron and SNN continuous inference proposed in Section \ref{section4.3}. Section \ref{section5.3} shows the ultra-high power efficiency of our SNN on FPGA.

 We select some videos from the MOT challenge \cite{leal2015motchallenge} as the validation dataset for our experiments. The spike datasets involved in this work are spiking streams in continuous scenes, which are captured using spiking cameras or by encoding the video with spike encoder \cite{hu2022optical,kim2020spiking}. 
 
The detection results of our experiments are evaluated using mAP50($\%$). The experiments are performed on Ubuntu system. Our simulation is based on the Pytorch framework and we conducted all experiments on NVIDIA Tesla V100 32G GPUs.

\subsection{Comparison of the Two SNNs}\label{section5.1}

\begin{figure}
\centering
  \subfigure[Accuracy after compression by different methods]{\includegraphics[width=0.48\textwidth]{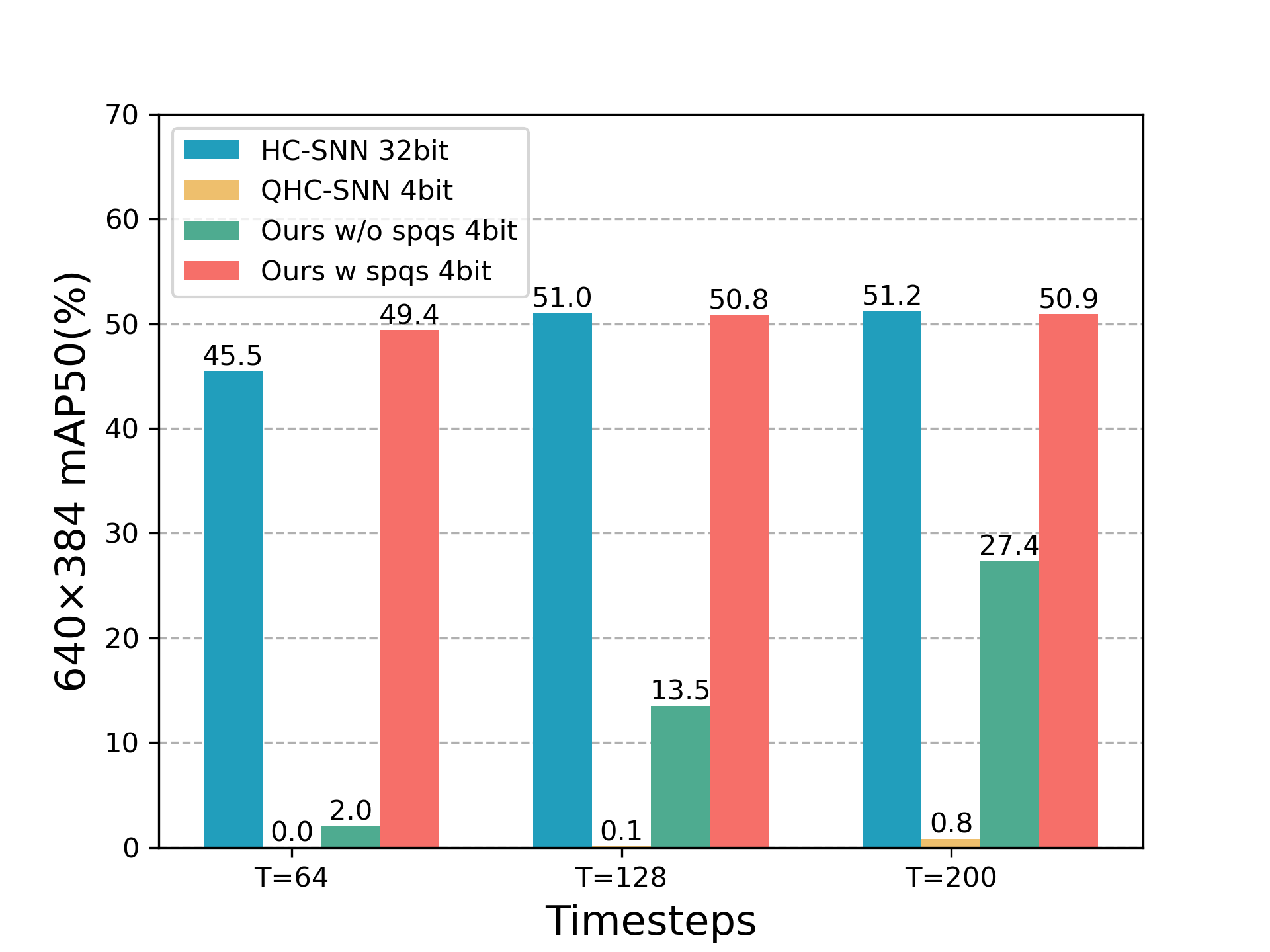}}
  \label{ST0}
  \subfigure[Performance of the two inference methods]{\includegraphics[width=0.48\textwidth]{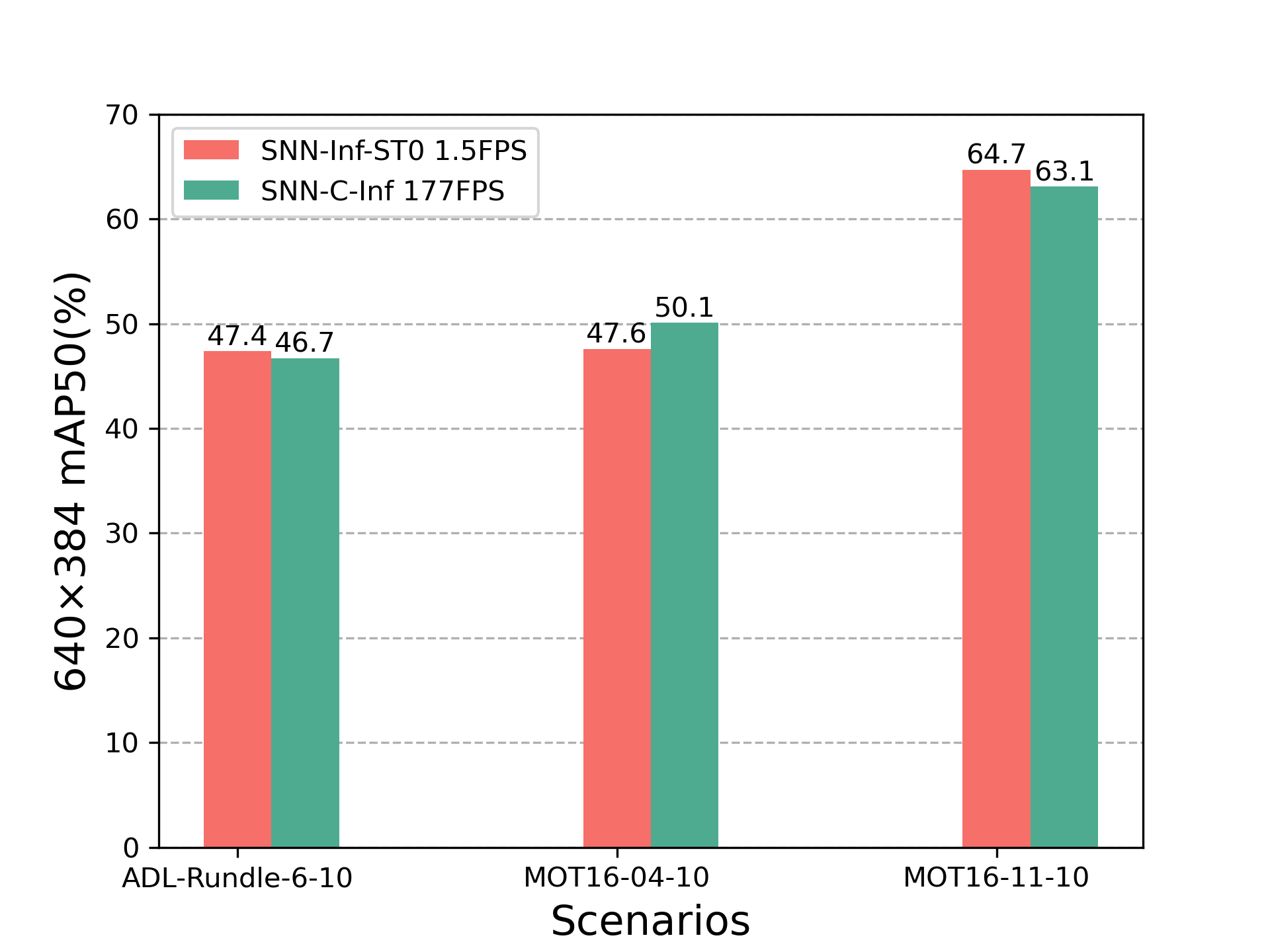}}
  \label{tab_directq}
  \caption{(a): Comparison of direct quantize the HC-SNN\cite{kim2020spiking} (QHC-SNN) and our methods (w or w/o spqs). (b):  Performance comparison of SNN-Inf-ST0 and SNN-C-Inf.}
  \label{runingtime}
  \vskip -0.1 in
\end{figure}

Since the previous method \cite{kim2020spiking} is not open source, we use an improved version (high complexity SNN (HC-SNN\cite{kim2020spiking})) to represent it.
The first experiment explores the performance between high complexity SNN (39.2MB) and low complexity SNN (1.5MB). High complexity SNN (HC-SNN\cite{kim2020spiking}) uses 32-bit floating point precision to store the weights, which is converted from high complexity ANN (HC-ANN\cite{kim2020spiking}) by using ANN to SNN conversion methods \cite{kim2020spiking,vaswani2017attention,ding2021optimal,li2022spike}. HC-ANN\cite{kim2020spiking} is trained according to the previous methods \cite{kim2020spiking} \cite{li2022spike} etc. Its network architecture is similar to that of tiny-yolov3. Our low complexity SNN (\textbf{Ours}) uses 4-bit integer precision to store the weights, which is converted from the initial compact quantized ANN (\textbf{QANN}) by using our method in Section \ref{section4.1}. We compare the size of the input data at 640$\times$384. Similarly we also compare in the cases of time steps T$=$64, T$=$128, T$=$200 and T$=$256 respectively. The Table \ref{tab:640map} shows the results when the input data size is 640$\times$384. In addition we tried to prune and directly quantize the HC-SNN\cite{kim2020spiking} (QHC-SNN) to reduce the bit-width of the weights, but there is a complete loss of performance. We also performed ablation experiments, whether to use our proposed scale-aware pseudo-quantization scheme(spqs) or not. The accuracy after compression by different methods is shown in Fig. \ref{runingtime}. We can summarize the following conclusions:

First, the results of HC-SNN\cite{kim2020spiking} show that our HC-SNN\cite{kim2020spiking} have good performance. Compared to the previous method \cite{kim2020spiking} which takes thousands of time steps, we need only 64 time steps for SNN object detection, even exceeding the performance of the original ANN in some scenarios. Second, compared to the direct quantification scheme, the performance of the low complexity SNN obtained by our method is much better than the QHC-SNN which has the same model size. And after using our proposed scale-aware pseudo-quantization scheme, the performance is almost lossless compared to the HC-SNN\cite{kim2020spiking}. The results show that the performance of our low complexity SNN is comparable to the initial compact quantized ANN, and even better than QANN in some scenarios. By carefully comparing the mAP50 in Table \ref{tab:640map}, we can find that our low complexity SNN can achieve very close performance to HC-SNN\cite{kim2020spiking} in all time step cases and all scenarios. However, our model size is only 1/26 of the original model size. This is an important reason why we can deploy it to FPGA.
\begin{table*}
\centering
\resizebox{0.98\textwidth}{15.8mm}{
\begin{tabular}{@{}c|cc|cc|cc|cc|cc@{}}
\toprule
\textbf{640$\times$384 mAP50($\%$)} & \multicolumn{2}{c|}{ANN}                                         & \multicolumn{2}{c|}{T=64}                                        & \multicolumn{2}{c|}{T=128}                                       & \multicolumn{2}{c|}{T=200}                                       & \multicolumn{2}{c}{T=256}                                       \\ \midrule
\textbf{Model size}         & \multicolumn{1}{l}{39.2MB} & \multicolumn{1}{l|}{\textbf{1.5MB}} & \multicolumn{1}{l}{39.2MB} & \multicolumn{1}{l|}{\textbf{1.5MB}} & \multicolumn{1}{l}{39.2MB} & \multicolumn{1}{l|}{\textbf{1.5MB}} & \multicolumn{1}{l}{39.2MB} & \multicolumn{1}{l|}{\textbf{1.5MB}} & \multicolumn{1}{l}{39.2MB} & \multicolumn{1}{l}{\textbf{1.5MB}} \\ \midrule
\textbf{Scenarios}          & HC-ANN                     & QANN                                & HC-SNN                     & \textbf{Ours}                       & HC-SNN                     & \textbf{Ours}                       & HC-SNN                     & \textbf{Ours}                       & HC-SNN                     & \textbf{Ours}                      \\ \midrule
ADL-Rundle-6                & 50.2                      & 49.8                               & 45.5                      & 49.4                               & 51.0                      & 50.8                               & 51.2                      & 50.9                               & 51.4                      & 50.9                              \\
ADL-Rundle-6\_gray          & 50.2                      & 49.8                               & 42.5                      & 49.4                               & 51.2                      & 50.7                               & 51.4                      & 50.5                               & 51.3                      & 50.7                              \\
MOT-11                      & 60.8                      & 62.2                               & 27.0                       & 59.7                               & 59.0                       & 61.5                               & 60.3                      & 61.9                               & 60.5                      & 62.0                              \\
AVG\_TownCentre             & 53.1                      & 50.4                               & 44.4                      & 47.8                               & 50.8                      & 50.2                               & 53.7                      & 50.2                               & 53.9                      & 50.4                              \\
ADL-Rundle-8                & 43.3                      & 40.9                               & 37.9                      & 37.0                               & 44.0                      & 38.4                               & 44.0                      & 38.4                               & 44.4                      & 38.5                              \\ \bottomrule
\end{tabular}}
\caption{Comparison of the object detection performance(640$\times$384 pixels mAP50) between our low complexity SNN (Ours) and the high complexity SNN (HC-SNN\cite{kim2020spiking}) in different scenarios.}
  \label{tab:640map}
  \vskip -0.2 in
\end{table*}

\subsection{Comparison of Different Inference Methods}\label{section5.2}





 

In this experiment, the input spiking data size is 640$\times$384 and the default time steps is T$=$200. Part of the experiment was tested on both SNN models (\textbf{Ours} and \textbf{HC-SNN\cite{kim2020spiking}}). Our experiment is set up with two types of inference. The first inference is the previous method using SNN inference (SNN-Inf-ST0\cite{cao2015spiking}) with IF neurons, the second is the proposed SNN continuous inference (\textbf{SNN-C-Inf}). 

Fig. \ref{runingtime} shows the performance comparison of SNN-Inf-ST0\cite{cao2015spiking} method and SNN-C-Inf method on our model. The experiment is to test the time consumed by using SNN inference and SNN inference continuous inference on GPU (32-bit floating-point inference), comparing the two methods running on the computer while outputting the same number of frames results. After conducting multiple tests in various scenarios, the average FPS for the two inference methods are as follows: SNN-C-INF: 177 and SNN-Inf-ST0\cite{cao2015spiking}: 1.5. Our approach has significant advantages, and at the same time our accuracy of SNN-C-INF inference can be almost equal to that of the SNN-Inf-ST0\cite{cao2015spiking}.

\subsection{Comparison of the Power Efficiency}\label{section5.3}

\begin{table*}

\centering
\resizebox{0.99\textwidth}{23.8mm}{
\begin{tabular}{@{}c|c|c|c|c|c@{}}
\toprule
\textbf{Method}                                      & Sim yolo v2\cite{redmon2017yolo9000}                  & Tiny yolo v2\cite{nguyen2019high}                 & Tiny yolo v3\cite{ahmad2020accelerating}                 & LeNet-SNN*\cite{ju2020fpga}                  & \textbf{Ours}    \\ \midrule
\textbf{Platform}                  & GPU         & FPGA         & FPGA         & FPGA      & FPGA            \\ \midrule
\textbf{Frequency}               & 1 GHz       & 200 MHz      & 200 MHz      & 150 MHz   & 150 MHz         \\
\textbf{GOP}                       & 6.5         & 2.64         & 2.1          & 0.22      & 1.4             \\
\textbf{Weight bit}                & 32          & 1            & 1            & 8         & 4               \\
\textbf{Activation bit}            & 32          & 6            & 18           & 1         & 1               \\
\textbf{Image Size}                & 256$\times$256     & 256$\times$256      & 256$\times$256      & 10$\times$28$\times$28  & 256$\times$256         \\
\textbf{FPS}                       & 232         & 176          & 219          & 164       & \textbf{681}    \\
\textbf{Throughput(GOPS)}          & 1512        & 464.7        & 460.8        & 36.08     & 954             \\
\textbf{Power(W)}                  & 170         & 8.7          & 4.81         & 4.6       & \textbf{2.437}  \\
\textbf{Power efficiency(GOP/s/W)} & 8.89        & 53.29        & 95.8         & 7.84      & \textbf{391.46} \\
\textbf{Power efficiency(FPS/W)}   & 1.365       & 20.23        & 45.53        & 35.65     & \textbf{279.44} \\ \bottomrule
\end{tabular}}
  \label{tab:Comparison in FPGA}
  \caption{Comparison of the power efficiency of GPU or FPGA deployments.}

\end{table*}

 To verify the performance of our highly efficient and fast SNN in the application, we deploy it into FPGA. Most weights of the deployed SNN network are stored using 4-bit. To the best of our knowledge, we are the first experiment to deploy a SNN for object detection to FPGA. Thus, we compared it to previous work on implementing ANN on GPU\cite{redmon2017yolo9000} or FPGA\cite{nguyen2019high,ahmad2020accelerating}. We also compared it with LeNet-SNN\cite{ju2020fpga} for classification tasks on FPGA. We set the input image resolutions of 256$\times$256. For method\cite{ju2020fpga}, their inference takes 10 time steps and the input is a 28$\times$28 MINST picture. We did not evaluate and compare performance due to the different tasks of the comparison methods\cite{redmon2017yolo9000,ju2020fpga,nguyen2019high,ahmad2020accelerating}. We compared the throughput (GOPS), power (W), and power efficiency (GOPS/W or FPS/W) of the devices. Table \ref{tab:Comparison in FPGA}
 shows the resources used and the power consumption achieved by each method. Experimental results show that we only need 2.437W of power consumption to achieve a detection speed of 681 FPS when the input image size is 256$\times$256. There is a huge improvement in power efficiency compared to the previous methods\cite{redmon2017yolo9000,ju2020fpga,nguyen2019high,ahmad2020accelerating}. If we set the input size as 224$\times$224, experimental results show that the detection speed will even be increased to more than 800FPS.

\section{Conclusion}
This paper is dedicated to the research of extremely efficient SNN to achieve super  high-speed inference. Specifically, we first generate an initial compact quantized ANN and convert it to a low complexity SNN, and then construct a SNN continuous inference scheme to realize high-speed object detection. Since there are almost no researches in these areas yet, we did our best to conduct the comparison with relevant experiments\cite{kim2020spiking,cao2015spiking,nguyen2019high,ahmad2020accelerating}. Our generation method compresses the model size 26$\times$ times and helps to restore model accuracy to near-identical levels as the original\cite{kim2020spiking} at the same time. In addition, our inference scheme helps to improve the information utilization and inference speed of SNN. For the application, we implement the first SNN for object detection on the FPGA platform. Beyond the object detection task, the proposed methods are theoretically generalizable to other SNN tasks.

\bibliography{egbib}

\begin{thebibliography}{29}
\providecommand{\natexlab}[1]{#1}
\providecommand{\url}[1]{\texttt{#1}}
\expandafter\ifx\csname urlstyle\endcsname\relax
  \providecommand{\doi}[1]{doi: #1}\else
  \providecommand{\doi}{doi: \begingroup \urlstyle{rm}\Url}\fi

\bibitem[Ahmad et~al.(2020)Ahmad, Pasha, and Raza]{ahmad2020accelerating}
Afzal Ahmad, Muhammad~Adeel Pasha, and Ghulam~Jilani Raza.
\newblock Accelerating tiny yolov3 using fpga-based hardware/software co-design.
\newblock In \emph{2020 IEEE International Symposium on Circuits and Systems (ISCAS)}, pages 1--5. IEEE, 2020.

\bibitem[Cao et~al.(2015)Cao, Chen, and Khosla]{cao2015spiking}
Yongqiang Cao, Yang Chen, and Deepak Khosla.
\newblock Spiking deep convolutional neural networks for energy-efficient object recognition.
\newblock \emph{International Journal of Computer Vision}, 113\penalty0 (1):\penalty0 54--66, 2015.

\bibitem[Diehl et~al.(2015)Diehl, Neil, Binas, Cook, Liu, and Pfeiffer]{diehl2015fast}
Peter~U Diehl, Daniel Neil, Jonathan Binas, Matthew Cook, Shih-Chii Liu, and Michael Pfeiffer.
\newblock Fast-classifying, high-accuracy spiking deep networks through weight and threshold balancing.
\newblock In \emph{2015 International joint conference on neural networks (IJCNN)}, pages 1--8. ieee, 2015.

\bibitem[Ding et~al.(2021)Ding, Yu, Tian, and Huang]{ding2021optimal}
Jianhao Ding, Zhaofei Yu, Yonghong Tian, and Tiejun Huang.
\newblock Optimal ann-snn conversion for fast and accurate inference in deep spiking neural networks.
\newblock \emph{arXiv preprint arXiv:2105.11654}, 2021.

\bibitem[Hu et~al.(2022)Hu, Zhao, Ding, Ma, Shi, Xiong, and Huang]{hu2022optical}
Liwen Hu, Rui Zhao, Ziluo Ding, Lei Ma, Boxin Shi, Ruiqin Xiong, and Tiejun Huang.
\newblock Optical flow estimation for spiking camera.
\newblock In \emph{Proceedings of the IEEE/CVF Conference on Computer Vision and Pattern Recognition}, pages 17844--17853, 2022.

\bibitem[Jacob et~al.(2018)Jacob, Kligys, Chen, Zhu, Tang, Howard, Adam, and Kalenichenko]{jacob2018quantization}
Benoit Jacob, Skirmantas Kligys, Bo~Chen, Menglong Zhu, Matthew Tang, Andrew Howard, Hartwig Adam, and Dmitry Kalenichenko.
\newblock Quantization and training of neural networks for efficient integer-arithmetic-only inference.
\newblock In \emph{Proceedings of the IEEE conference on computer vision and pattern recognition}, pages 2704--2713, 2018.

\bibitem[Ju et~al.(2020)Ju, Fang, Yan, Xu, and Tang]{ju2020fpga}
Xiping Ju, Biao Fang, Rui Yan, Xiaoliang Xu, and Huajin Tang.
\newblock An fpga implementation of deep spiking neural networks for low-power and fast classification.
\newblock \emph{Neural computation}, 32\penalty0 (1):\penalty0 182--204, 2020.

\bibitem[Kim et~al.(2020)Kim, Park, Na, and Yoon]{kim2020spiking}
Seijoon Kim, Seongsik Park, Byunggook Na, and Sungroh Yoon.
\newblock Spiking-yolo: spiking neural network for energy-efficient object detection.
\newblock In \emph{Proceedings of the AAAI conference on artificial intelligence}, volume~34, pages 11270--11277, 2020.

\bibitem[Kim et~al.(2022)Kim, Li, Park, Venkatesha, Yin, and Panda]{kim2022exploring}
Youngeun Kim, Yuhang Li, Hyoungseob Park, Yeshwanth Venkatesha, Ruokai Yin, and Priyadarshini Panda.
\newblock Exploring lottery ticket hypothesis in spiking neural networks.
\newblock In \emph{European Conference on Computer Vision}, pages 102--120. Springer, 2022.

\bibitem[Krishnamoorthi(2018)]{krishnamoorthi2018quantizing}
Raghuraman Krishnamoorthi.
\newblock Quantizing deep convolutional networks for efficient inference: A whitepaper.
\newblock \emph{arXiv preprint arXiv:1806.08342}, 2018.

\bibitem[Leal-Taix{\'e} et~al.(2015)Leal-Taix{\'e}, Milan, Reid, Roth, and Schindler]{leal2015motchallenge}
Laura Leal-Taix{\'e}, Anton Milan, Ian Reid, Stefan Roth, and Konrad Schindler.
\newblock Motchallenge 2015: Towards a benchmark for multi-target tracking.
\newblock \emph{arXiv preprint arXiv:1504.01942}, 2015.

\bibitem[Li et~al.(2022)Li, He, Dong, Kong, and Zeng]{li2022spike}
Yang Li, Xiang He, Yiting Dong, Qingqun Kong, and Yi~Zeng.
\newblock Spike calibration: Fast and accurate conversion of spiking neural network for object detection and segmentation.
\newblock \emph{arXiv preprint arXiv:2207.02702}, 2022.

\bibitem[Liu et~al.(2022)Liu, Zhao, Chen, Wang, and Dai]{liu2022dynsnn}
Fangxin Liu, Wenbo Zhao, Yongbiao Chen, Zongwu Wang, and Fei Dai.
\newblock Dynsnn: A dynamic approach to reduce redundancy in spiking neural networks.
\newblock In \emph{ICASSP 2022-2022 IEEE International Conference on Acoustics, Speech and Signal Processing (ICASSP)}, pages 2130--2134. IEEE, 2022.

\bibitem[Liu et~al.(2017)Liu, Li, Shen, Huang, Yan, and Zhang]{liu2017learning}
Zhuang Liu, Jianguo Li, Zhiqiang Shen, Gao Huang, Shoumeng Yan, and Changshui Zhang.
\newblock Learning efficient convolutional networks through network slimming.
\newblock In \emph{Proceedings of the IEEE international conference on computer vision}, pages 2736--2744, 2017.

\bibitem[Nguyen et~al.(2019)Nguyen, Nguyen, Kim, and Lee]{nguyen2019high}
Duy~Thanh Nguyen, Tuan~Nghia Nguyen, Hyun Kim, and Hyuk-Jae Lee.
\newblock A high-throughput and power-efficient fpga implementation of yolo cnn for object detection.
\newblock \emph{IEEE Transactions on Very Large Scale Integration (VLSI) Systems}, 27\penalty0 (8):\penalty0 1861--1873, 2019.

\bibitem[Panchapakesan et~al.(2022)Panchapakesan, Fang, and Li]{panchapakesan2022syncnn}
Sathish Panchapakesan, Zhenman Fang, and Jian Li.
\newblock Syncnn: Evaluating and accelerating spiking neural networks on fpgas.
\newblock \emph{ACM Transactions on Reconfigurable Technology and Systems (TRETS)}, 2022.

\bibitem[Park et~al.(2019)Park, Kim, Choe, and Yoon]{park2019fast}
Seongsik Park, Seijoon Kim, Hyeokjun Choe, and Sungroh Yoon.
\newblock Fast and efficient information transmission with burst spikes in deep spiking neural networks.
\newblock In \emph{2019 56th ACM/IEEE Design Automation Conference (DAC)}, pages 1--6. IEEE, 2019.

\bibitem[Redmon and Farhadi(2017)]{redmon2017yolo9000}
Joseph Redmon and Ali Farhadi.
\newblock Yolo9000: better, faster, stronger.
\newblock In \emph{Proceedings of the IEEE conference on computer vision and pattern recognition}, pages 7263--7271, 2017.

\bibitem[Redmon et~al.(2016)Redmon, Divvala, Girshick, and Farhadi]{redmon2016you}
Joseph Redmon, Santosh Divvala, Ross Girshick, and Ali Farhadi.
\newblock You only look once: Unified, real-time object detection.
\newblock In \emph{Proceedings of the IEEE conference on computer vision and pattern recognition}, pages 779--788, 2016.

\bibitem[Roy et~al.(2019)Roy, Jaiswal, and Panda]{roy2019towards}
Kaushik Roy, Akhilesh Jaiswal, and Priyadarshini Panda.
\newblock Towards spike-based machine intelligence with neuromorphic computing.
\newblock \emph{Nature}, 575\penalty0 (7784):\penalty0 607--617, 2019.

\bibitem[Rueckauer et~al.(2017)Rueckauer, Lungu, Hu, Pfeiffer, and Liu]{rueckauer2017conversion}
Bodo Rueckauer, Iulia-Alexandra Lungu, Yuhuang Hu, Michael Pfeiffer, and Shih-Chii Liu.
\newblock Conversion of continuous-valued deep networks to efficient event-driven networks for image classification.
\newblock \emph{Frontiers in neuroscience}, 11:\penalty0 682, 2017.

\bibitem[Sengupta et~al.(2019)Sengupta, Ye, Wang, Liu, and Roy]{sengupta2019going}
Abhronil Sengupta, Yuting Ye, Robert Wang, Chiao Liu, and Kaushik Roy.
\newblock Going deeper in spiking neural networks: Vgg and residual architectures.
\newblock \emph{Frontiers in neuroscience}, 13:\penalty0 95, 2019.

\bibitem[Shrestha and Orchard(2018)]{shrestha2018slayer}
Sumit~B Shrestha and Garrick Orchard.
\newblock Slayer: Spike layer error reassignment in time.
\newblock \emph{Advances in neural information processing systems}, 31, 2018.

\bibitem[Tavanaei and Maida(2019)]{tavanaei2019bp}
Amirhossein Tavanaei and Anthony Maida.
\newblock Bp-stdp: Approximating backpropagation using spike timing dependent plasticity.
\newblock \emph{Neurocomputing}, 330:\penalty0 39--47, 2019.

\bibitem[Vaswani et~al.(2017)Vaswani, Shazeer, Parmar, Uszkoreit, Jones, Gomez, Kaiser, and Polosukhin]{vaswani2017attention}
Ashish Vaswani, Noam Shazeer, Niki Parmar, Jakob Uszkoreit, Llion Jones, Aidan~N Gomez, {\L}ukasz Kaiser, and Illia Polosukhin.
\newblock Attention is all you need.
\newblock \emph{Advances in neural information processing systems}, 30, 2017.

\bibitem[Wang et~al.(2022{\natexlab{a}})Wang, Zhang, Chen, and Qu]{wang2022signed}
Yuchen Wang, Malu Zhang, Yi~Chen, and Hong Qu.
\newblock Signed neuron with memory: Towards simple, accurate and high-efficient ann-snn conversion.
\newblock In \emph{International Joint Conference on Artificial Intelligence}, 2022{\natexlab{a}}.

\bibitem[Wang et~al.(2022{\natexlab{b}})Wang, Gu, Goh, Zhou, and Luo]{wang2022efficient}
Zhehui Wang, Xiaozhe Gu, Rick Siow~Mong Goh, Joey~Tianyi Zhou, and Tao Luo.
\newblock Efficient spiking neural networks with radix encoding.
\newblock \emph{IEEE Transactions on Neural Networks and Learning Systems}, 2022{\natexlab{b}}.

\bibitem[Wu et~al.(2019)Wu, Deng, Li, Zhu, Xie, and Shi]{wu2019direct}
Yujie Wu, Lei Deng, Guoqi Li, Jun Zhu, Yuan Xie, and Luping Shi.
\newblock Direct training for spiking neural networks: Faster, larger, better.
\newblock In \emph{Proceedings of the AAAI Conference on Artificial Intelligence}, volume~33, pages 1311--1318, 2019.

\bibitem[Ye et~al.(2018)Ye, Lu, Lin, and Wang]{ye2018rethinking}
Jianbo Ye, Xin Lu, Zhe Lin, and James~Z Wang.
\newblock Rethinking the smaller-norm-less-informative assumption in channel pruning of convolution layers.
\newblock \emph{arXiv preprint arXiv:1802.00124}, 2018.

\end{thebibliography}
\end{document}